\documentclass{article}

\usepackage{ifthen}
\newboolean{neurips_version}
\setboolean{neurips_version}{true}
\ifthenelse{\boolean{neurips_version}}
{\PassOptionsToPackage{numbers, compress}{natbib}
 \usepackage[preprint]{neurips_2022}
}
{
\usepackage{PRIMEarxiv}
 \usepackage[numbers]{natbib}
}

\usepackage[utf8]{inputenc} 
\usepackage[T1]{fontenc}    
\usepackage{hyperref}       
\usepackage{url}            
\usepackage{booktabs}       
\usepackage{amsfonts,amsmath,amsthm,amssymb}       
\usepackage{nicefrac}       
\usepackage{microtype}      
\usepackage{xcolor}         
\usepackage[disable]{todonotes}
\usepackage{enumitem}
\usepackage{caption}

\usepackage{mathtools} 
\usepackage{tikz}
\usetikzlibrary{calc}
\usepackage{pgfplots}
\usepackage{graphicx}

\usepackage{amsmath}
\usepackage{algorithm, algpseudocode}
\usepackage{amsfonts}  

\usepackage[english]{babel}
\usepackage{amsthm}

\usepackage{subfig}
\usepackage[export]{adjustbox}

\usepackage{hhline}
\usepackage{makecell, caption, booktabs}
\usepackage{siunitx}

\usepackage{multirow}

\usepackage{amsmath, nccmath}
\usepackage{bigstrut}

\usepackage{booktabs}

\usepackage{lipsum}
\graphicspath{{media/}}     

\usepackage{CJKutf8}
\usepackage[framed,numbered,autolinebreaks,useliterate]{mcode}

\title{ Activation Functions Not To Active: A Plausible Theory on Interpreting Neural Networks }

\author{ \href{https://orcid.org/0000-0003-0378-0607}{\includegraphics[scale=0.06]{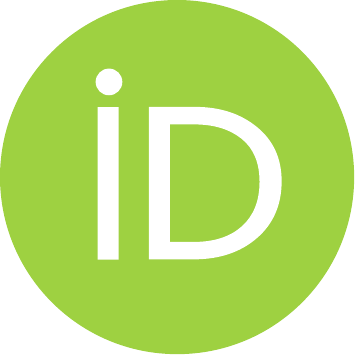}\hspace{1mm}John Chiang} \\                             
                                      \\
	\texttt{john.chiang.smith@gmail.com} 
}

\date{}

\theoremstyle{remark}

\renewcommand{\epsilon}{\varepsilon}

\makeatletter
\def\namedlabel#1#2{\begingroup
    #2%
    \def\@currentlabel{#2}%
    \phantomsection\label{#1}\endgroup
}
\makeatother

\algnewcommand{\LeftComment}[1]{\Statex \(\triangleright\) #1}
\algnewcommand{\LineCommentStep}[1]{\Statex \textbf{[Step #1]:} }
\makeatletter
\newlength{\trianglerightwidth}
\algnewcommand{\LineComment}[1]{\Statex \hskip\ALG@thistlm $\triangleright$ #1}
\algnewcommand{\LineCommentCont}[1]{\Statex \hskip\ALG@thistlm%
  \parbox[t]{\dimexpr\linewidth-\ALG@thistlm}{\hangindent=\trianglerightwidth \hangafter=1 \strut$\triangleright$ #1\strut}}
\algnewcommand{\LeftLineCommentCont}[1]{\Statex \hskip\ALG@thistlm%
  \parbox[t]{\dimexpr\linewidth-\ALG@thistlm}{\leftskip=\algorithmicindent \hangindent=\trianglerightwidth \hangafter=1 \strut$\triangleright$ #1\strut}}

\newcommand{\mysplit}[1]{%
  \begin{tabular}{@{}c@{}}
    #1
  \end{tabular}
  }
  
\begin{document}

\maketitle

\begin{abstract}%
Researchers commonly believe that neural networks model a high-dimensional space but cannot give a clear definition of this space. What is this space? What is its dimension? And does it has finite dimensions? 
In this paper, we develop a plausible theory on interpreting neural networks in terms of the role of activation functions in neural networks and define a high-dimensional (more precisely, an infinite-dimensional) space that neural networks including deep-learning networks could create. We show that the activation function acts as a magnifying function that maps the low-dimensional linear space into an infinite-dimensional space, which can distinctly identify the polynomial approximation of any multivariate continuous function of the variable values being the same features of the given dataset.

Given a dataset with each example of $d$ features $f_1$, $f_2$, $\cdots$, $f_d$, we believe that neural networks model a  special space with infinite dimensions, each of which is a monomial $$\prod_{i_1, i_2, \cdots, i_d} f_1^{i_1} f_2^{i_2} \cdots f_d^{i_d}$$ for some non-negative integers  ${i_1, i_2, \cdots, i_d}  \in  \mathbb{Z}_{0}^{+}=\{0,1,2,3,\ldots\} $. We term such an infinite-dimensional space a $\textit{ Super Space (SS)}$. We see such a dimension as the minimum information unit. Every neuron node previously through an activation layer in neural networks is a $\textit{ Super Plane (SP) }$, which is actually a polynomial of infinite degree. 

This $\textit{ Super Space }$ is something like a coordinate system, in which every multivalue function can be represented by a $\textit{ Super Plane }$.
From this perspective, a neural network for regression tasks can be seen as an extension of linear regression, i.e. an advanced variant of linear regression with infinite-dimensional features. 

We also show that training NNs could at least be reduced to solving a system of nonlinear equations. 

\todo{Show precise somewhere that our results are new in the iid setting as well.}

\end{abstract}

\listoftodos

\section{Introduction}

\subsection{ Background }
Neural Networks (NN), including the currently popular deep-learning ones, are  playing a more and more significant role in important real-world domains, such as image classification, face recognition, and machine translation. 
Despite the great success of neural networks over the past decade, NNs are routinely considered a ``black box'' due to the lack of interpretability, leaving the big questions about how they arrive at predictions or decisions and why they work and have such a great performance.

\subsection{ Previous Work }

Universal Approximation Theorem~\cite{hornik1989multilayer} and Stone-Weierstrass Theorem~\cite{katznelson1961stone} show that any continuous and bounded function can be approximated by NNs and polynomials, respectively.  It might at first seem that our work here merely confirms this fact. However, we demonstrate a more subtle but much closer connection than that.  We believe that the NN actually models an infinite-dimensional space, which we term $\textit{Super Space (SS)}$. Any multivalue polynomial to approximate some multivalue function can be represented in this special space by a hyperplane with infinite dimensions, which we name  $\textit{Super Plane (SP)}$. We are interested in both the NN training and inference processes themselves; we show that training NNs is to search a proper SP to represent a polynomial approximation and that NN inference is just to compute the outputs of several finite-dimensional hyperplanes to approximate some SPs. Just like calculating a limited sum for the Taylor series in real-world practice, we could approximate this infinite-dimensional space and planes by a hyperspace and some hyperplanes both with finite dimensions, respectively, in order to analyze and use this super space and super planes.  This special space can and should be approximated in real-world applications by a hyperspace with finite dimensions, whose number depends on how many outputs there are in the last layer.  We point out here that such an SS approximation is something like a Cartesian coordinate system and that such an SP approximation is a multivalue polynomial. 

Our work is primarily on general feedforward NNs. However, we believe our ideas could be adapted to specialized networks such as convolutional NNs (CNNs) and recurrent NNs and so on. For instance, we view the convolutional and pooling layers in CNNs as largely playing the role of manipulating several SPs together, conducted for feature selection to be carried in polynomials approximation of some planes. 

It may well be that such an infinite-dimensional space has relationships with polynomial regression~\cite{morala2021towards, cheng2018polynomial}, support vector machines, and so on. However, this work aims not to explore the possible relationship between this special space and other methods, nor does it aim to compare other methods to SS and NN in various performances. There is no implied claim that SS outperforms other machine learning methods.  Instead, we claim that NN $\leftrightarrow$ SS implies a very close direct relationship between SS and NNs, and explore the consequences and possible applications based on this close relationship.

\subsection{ Contributions }
The present work will make the following contributions:

\begin{enumerate}[label=(\alph*)]
	\item \label{PointA} We will show that in any NN, at each neuron with an activation function, there is a close correspondence to an infinite-dimensional hyperspace (a Super Space); in essence, NNs, including the popular form of many-layered deep learning networks, model a unique SS that can represent every multivariate polynomial of some certain variable values, and the output of such a neuron represents a  Super Plane (SP). We refer to this clear correspondence here as NN $\leftrightarrow$ SS.  

	\item An important aspect of NN $\leftrightarrow$ SS is that the Super Space has infinite dimensions, each of which is a monomial. The SP in this space is like the line $ y = x + 2 $ in the $\textit{ X-Y} $ Cartesian coordinate system. In other words, our findings could be interpreted as saying that the end result of an NN can be represented by some coordinate system, except that this system has infinite dimensions.

	\item We exploit NN $\leftrightarrow$ SS in order to learn about the general properties of NNs via the basic knowledge of the properties of ordinary coordinate systems, such as the two-dimensional Cartesian system and the three-dimensional Cartesian system. This would turn out to provide new insights into aspects such as how to deal with categorical data.

	\item Property~\ref{PointA} suggests that in many applications, one might simply fit a super-plane approximation (namely, approximated by a polynomial) in the first place, bypassing NNs. This would overcome the disadvantage of selecting various architecture, choosing numerous hyper-parameters, and so on.
\end{enumerate}

Point~\ref{PointA} is the core idea behind this work, as it shows a much tight connection of NNs to SS that is never reported. The output of every last-layer neuron and hidden-layer neurons previously through activation layers is a Super Plane, something like a polynomial but not exactly a polynomial due to its infinite degree. Literatures~\cite{hornik1989multilayer} have noted some theoretical connections between NNs and polynomials but our contributions go much deeper. It shows that in essence,  conventional NNs and other deep NNs actually model an infinite-dimensional hyperspace and that in fact, the outputs of NNs are multiple infinite-dimensional hyperplanes but could be approximated by some finite ones, namely polynomials of a certain degree. Our focus will be on the activation function. Using a formal mathematical analysis on any activation function, we show $\textit{ why NNs are essentially a form of Super Space }$.  

Our work not only can be on general feedforward NNs but also can involve on specialized networks such as convolutional NNs (CNNs), recurrent NNs, and so on. We see NNs as a special space with infinite dimensions and their outputs as some special planes with infinite dimensions, rather than regarding NNs as polynomial regression. In real-world applications, such spaces and planes will be respectively replaced by hyperspaces and hyperplanes with finite dimensions.
In these cases, we have:
If the activation function is a polynomial of infinite degree (unlimited series) or is implemented by such one, an NN exactly models an infinite-dimensional space; otherwise, if the activation function is a polynomial of a certain degree, an NN still models an infinite-dimensional space but only with finite non-zero coefficient dimensions---the other dimensions simply all have zero coefficients.


\section{Methodology}

In this paper, the activation function only means nonlinear modern popular functions, such as sigmoid, ReLU, and tanh functions, which cannot be represented by $\Phi(x) = a\cdot x + b$  for some floating-point numbers $a$ and $b$.  Furthermore, such activation functions do not include polynomial activation functions with a certain degree but could be a polynomial of an infinite degree.

\subsection{Polynomial Approximation of Any Activation Function}
Taylor polynomials used to be commonly adopted to approximate activation functions, such as the sigmoid function,  at some points by calculating the function's derivatives. However, they are not a good candidate for approximating a whole function over a large range because they only provide a local approximation near a certain point. Taking into account the overall behavior of the function and not just the behavior near a specific point, the least-squares approximation, as a global approximation method minimizing the mean squared error, is more suitable for approximating a function over a large range. 


We develop a new approach to approximate any activation function, which can be briefly described as two steps. 
Given an activation function $\Phi(x)$:
 (1) we use the Fourier series to fit the function $\Phi(x)$ over the prescribed range, resulting in a Fourier expansion consisting of only the sine and cosine functions; and (2) we adopt the Taylor series to approximate the trigonometric functions at some points in the Fourier expansion from the first step, obtaining the desired polynomial approximation.

Fourier series is an expansion of a periodic function f(x) in terms of an infinite sum of sines and cosines. If a function $f(x)$ has a finite number of jump discontinuities, which is piecewise smooth~\cite{tolstov2012fourier}, then $f(x)$ has a Fourier series. Such a function is called to be piecewise smooth. For a function $f(x)$ of period $2l$, we can form its Fourier series $f(x) \sim \frac{1}{2}a_0 + \sum_{n=1}^{\infty}(a_n\cos{\frac{n\pi}{l}x} + b_n\sin{\frac{n\pi}{l}x})$, where $a_n = \frac{1}{l}\int_{-l}^lf(x)\cos{\frac{n\pi}{l}x}dx, (n=0,1,2,\ldots)$ and
$b_n = \frac{1}{l}\int_{-l}^lf(x)\sin{\frac{n\pi}{l}x}dx, (n=1,2,\ldots)$. 
The sum of its Fourier series will converge to $f(x)$ at the points of continuity and to the arithmetic mean of the right-hand and left-hand limits at the points of discontinuity. In real-world applications, it is usually a must to replace the infinite series $\sum_{n=1}^{\infty}$ with a finite one $\sum_{n=1}^{N}$ to approximate  the original function, where $N$ is a constant.




For a function defined only over a limited interval, like the sigmoid function over the domain $[-8, 8]$, 
it can be extended from its original domain onto the whole X-axis to be a periodic function. We then expand this periodic function in a  trigonometric series, which converges to $f(x)$ as long as it is a  piecewise smooth function. In this case, the sigmoid function over the domain $[-l, l]$ has such a Fourier series as $f(x) \sim \frac{1}{2}a_0 + \sum_{n=1}^{\infty}(a_n\cos{\frac{n\pi}{l}x} + b_n\sin{\frac{n\pi}{l}x}) = \frac{1}{2} + \sum_{n=1}^{k}( b_n\sin{\frac{n\pi}{l}x})$, where k is the degree for the trigonometric polynomial, $a_0$ is $0.5$, and $a_n = 0$.


As we mentioned before, Taylor series is  suitable for finding the value of a function $f(x)$ at a certain point $x_0$, but not for fitting the function itself over a wide range. For an infinite differentiable function $f(x)$, the Taylor series of the function $f(x)$ at $a$ (or centered at $a$) is: $f(x) = \sum_{n=0}^{\infty} \frac{f^{(n)}(a)}{n!}{(x-a)}^n$, where $f^{(n)}(a)$ is the $n$-$th$ derivative of $f(x)$ evaluated at the point $a$. The Taylor series with a special case $a = 0$ is given the name Maclaurin series. 
The Maclaurin series for $\sin x$ and $\cos x$ are  respectively  $\sin x = \sum_{n=0}^{\infty}{(-1)}^n\frac{x^{2n+1}}{(2n+1)!}$ and $\sin x = \sum_{n=0}^{\infty}{(-1)}^n\frac{x^{2n+1}}{(2n+1)!}$ for all real x. For practical use, a limited sum $\sum_{n=1}^{K}$ is enough to calculate the value of the function at some point, where $K$ is a constant that depends on how much accuracy we want. We then substitute these two polynomials into the finite Fourier series expansion from the first step, obtaining the final polynomial approximation of the activation function over a certain domain.  


It is straightforward to come to the conclusion that $\textbf{ to approximate any activation function over the }$ $\textbf{  whole real domain we need an infinite series polynomial }$  and that $\textbf{any (multivariate) function}$    $\textbf{ with a Fourier series can be approximated to arbitrary precision by a (multivariate) polynomial }$.

Based on the above observation, we make an assumption:  $\textbf{ Polynomial of an infinite degree to}$ $\textbf{ properly approximate some activation function is equal to  the activation function itself  }$, since the difference at any point between the activation function and the polynomial approximation can be arbitrarily small.
\subsection{ The Role of Activation Functions }
Activation functions play a crucial role in the success of neural networks and are an essential component of neural network design. Without activation functions, neural networks are just linear transformations of input data, which would limit their ability to model complex phenomena.

The traditional view of the role of an activation function in a neural network is that it determines whether a neuron should be activated or not. The main purpose of activation functions is to introduce nonlinearity into the output of neurons in neural networks. This nonlinearity is crucial for networks to model complex phenomena and learn nonlinear relationships between inputs and outputs.

\begin{figure}[htp]
\centering
\includegraphics[scale=0.6]{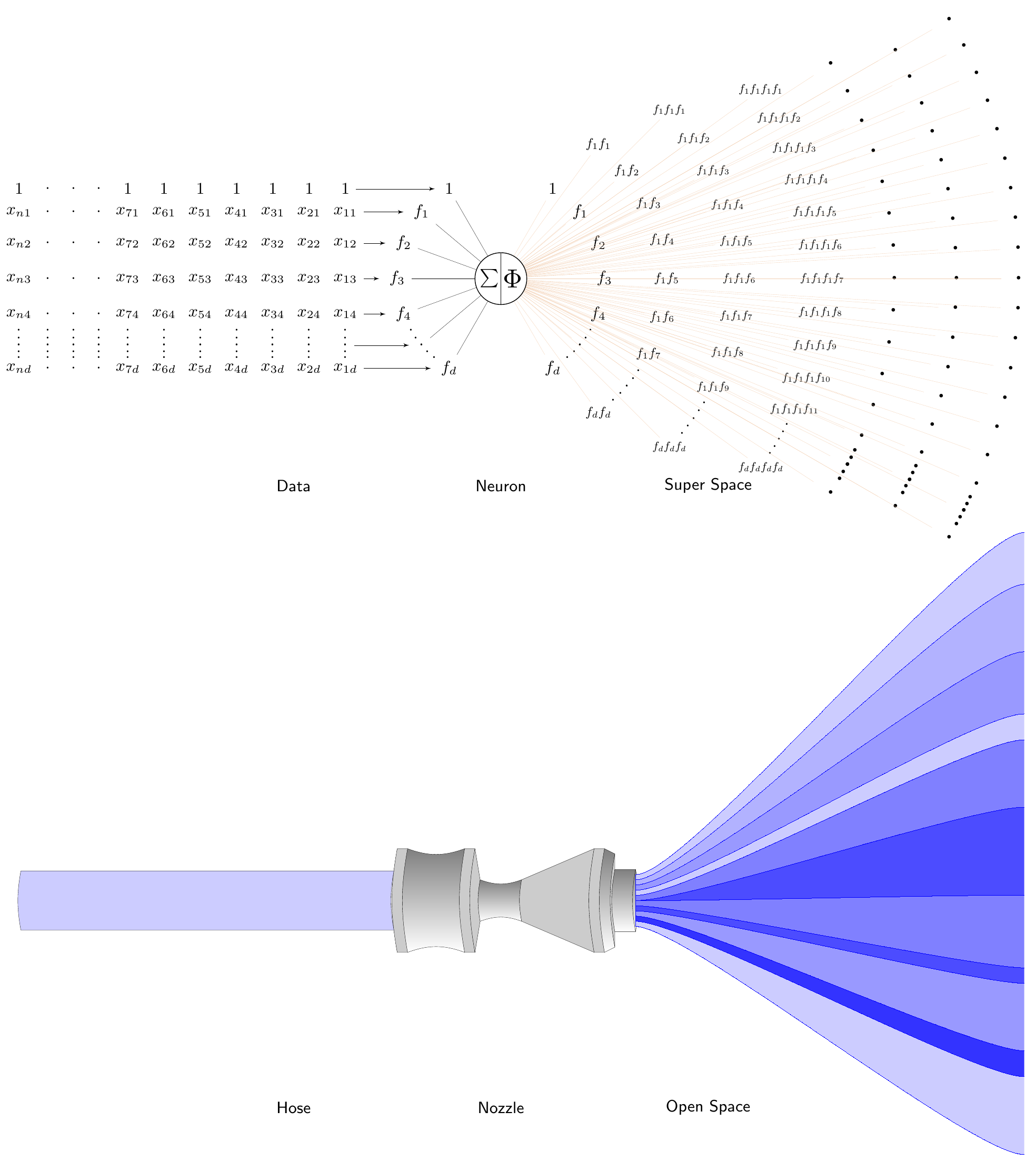}
\caption{
 The function of an activation function is just like that of a nozzle }
\label{ the role of an activation function }
\end{figure}

We present a new point of sight to enhance the traditional view: the activation function merely acts like activating a function or ``firing a cell'' but actually it plays as a magnifying function to elevate the original linear space onto a high-dimensional space with infinite dimensions, thereby finally leading to introduce non-linearity. It behaves like a nozzle squeezing the water stream into an open wide space. The different types of nozzle and the director it points to decide which open space the water would spray into. In a similar way, the activation function projects the input data concatenated together in a linear combination of several super planes into an infinite-dimensional space, each dimension of which is a minimum information unit. The different weight of each dimension on the input SP of the activation function will affect that of each dimension on the output SP. The role of an activation function whose input range is the whole real is as shown in Figure~\ref{ the role of an activation function }.

\subsection{The Sampling Procedure}
As long as the behavior of a multivalue function is observed, we need the sampling procedure to produce a data stream or dataset. This involves precision loss and would turn functions that cannot have a Fourier series into functions with one. The sampling procedure would fail to capture the true function and return a close function approximation that can have a Fourier series and that thus has a polynomial approximation.

\subsection{A Plausible Theory}
Consider a sample dataset $X \in \mathbb{R}^{n \times (1 + d)}$ consisting of $n$ observations of $d$ features $f_ 0 (= 1)$, $f_ 1$, $\cdots$, $f_ d$:

\begin{align*}
  X = 
 \begin{bmatrix}
 x_{10}    &   x_{11}   &  \cdots  & x_{1d}   \\
 x_{20}    &   x_{21}   &  \cdots  & x_{2d}   \\
 \vdots    &   \vdots   &  \ddots  & \vdots   \\
 x_{n0}    &   x_{n1}   &  \cdots  & x_{nd}   \\
 \end{bmatrix},     
 Y =  
 \begin{bmatrix}
 y_{1}     \\
 y_{2}     \\
 \vdots         \\
 y_{n}     \\
 \end{bmatrix} .
\end{align*}

Let the activation function $\Phi(x)$  be any modern one that can be approximated by, or in other words be equivariant to, a polynomial of infinite degree: $\Phi(x) = \sum_{i=0}^{\infty} c_i \cdot x^{i}$ for some real numbers $c_i$. 

Take the simple case as in Figure~\ref{ the role of an activation function } where there is only one neuron node.  The input to the node, including from the ``$1$'' node ($f_0$), will then be of the form $a_{0} + a_{01}f_1 + \cdots + a_{0d}f_d$ and the output to this node would be  $\sum_{i=0}^{\infty} c_i \cdot {(a_{0} + a_{01}f_1 + \cdots + a_{0d}f_d)}^{i}$, which is a polynomial of infinite degree.

We here develop a new theory that NNs model an infinite-dimensional space with each dimension as a monomial $$\prod_{i_1, i_2, \cdots, i_d} f_1^{i_1} f_2^{i_2} \cdots f_d^{i_d}$$ for some non-negative integers  ${i_1, i_2, \cdots, i_d}  \in  \mathbb{Z}_{0}^{+}=\{0,1,2,3,\ldots\} $. The output of each node that as long as comes through an activation function  in its previous layers  
is just an infinite-dimensional plane, 
namely a polynomial of infinite degree. We term such an infinite-dimensional space a $\textit{Super Space (SS)}$ and such an infinite-dimensional plane a $\textit{Super Plane (SP)}$. 

Since the real-world dataset has limited precision and a certain size, this SS can be approximated by a high-dimensional hyperspace with a limited number of dimensions.

Based on this theory, neural network inference is to compute the output of one or multiple such SPs, just like calculating the value $y(x = \textit{ some constant value } x_0 ) = a \cdot x + b$ in the two-dimensional coordinate system. To be specific, NNs compute the coordinates in the SS and calculate the output of the corresponding SP. In this sense, NNs can be seen as an extension of linear regression, just an advanced one. 
It seems that the training process aims to search important dimensions (minimum information unit) from the whole SS and decides which SP could best represent the multivalue function to fit in this mission.


According to this theory, NNs are not like some machine learning algorithms already exist.

\section{Applications}
We could analyze this infinite-dimensional space by studying ordinary coordinate systems like the two-dimensional coordinate system and the three-dimensional coordinate system.

\subsection{One-Dot Neural Networks}
From this perspective of this infinite-dimensional space, we could design a new NN architecture to solve the difficulty of handling categorical data in NNs. The current popular method is mainly by one-hot encoding, which will introduce more parameters into the NN. We here present a new novel NN architect that might crack this issue, which we term one-dot NN. 
We would like to leave the experiments testifying this new NN architecture as an open future work.

\subsection{Activation Functions Make Count}
The activation function actually plays an important role in NNs, to what extent important depending on the specific mission.  A well-known example is the invention of the activation function ReLU, which is commonly the first go-to choice made by most researchers.  Also in the above example with only one-node NN, various activation functions behave significantly differently. To classify a problem involving in period function, like judging the input number odd or even, a non-period function like ReLU or sigmoid function would perform badly. A one-hidden layer NN with non-period activation functions might only be able to approximate the odd-or-even problem over a certain range. On the other hand, a NN containing one period function like a sine or cosine function would perfectly approximate this problem over the whole integer domain. The complex NN architecture and numerous neuro nodes reduce the important effects of a certain activation function, but that doesn't mean activation functions don't matter.

\subsection{Both Classification and Regression}
Unlike most machine learning algorithms, NNs can be applied to both classification and regression problems, and have competitive performance in both cases. For simplicity, we assume that the regression problem involves making a single prediction. In this case, we could see the output of the SP as the only prediction. New coming examples would either  be mapped onto or close to this SP, resulting in a good prediction, or they will be far away from this SP, in which case we wouldn't expect good results.

For classification tasks, if the NN has multiple outputs in the last layer, then there are multiple class labels to consider. We can treat each output as a separate SP that measures the negative distance or difference between the position of the new example in the SS and the corresponding SP. 
The decision on which class to assign the new example made based on the outputs of the NN using the softmax function is to select the SP that is nearest to the new data's coordinates in the SS.
\subsection{Neural Network Model Compression}
Supposing that there is a already well-trained NN or even a CNN with a large number of parameters. We can actually calculate its SPs by replacing the activation functions with a perfect polynomial approximation, and then obtain the polynomial approximation $s$ for each SP. We then select a simple NN architecture with only one or two hidden layers and a smaller number of neuron nodes and also calculate its polynomial approximation $t$  for each output SP. Make sure the two polynomial approximations $s$ and $t$ have the same degrees for convenience in the calculation. The method of undetermined coefficients in mathematics can be used to determine each coefficient of polynomial approximation $t$. Once we get the polynomial $t$, we get the simpler NN with fewer parameters that share the same calculation circuit. In this way, we can compress a well-trained CNN into any NN architect we desire.

Note that in the inference phase, the drop-out technique usually wouldn't be adopted. Other techniques such as the batch normalization layer and maxing pooling layer are still manipulating polynomials to generate a new polynomial. For example, the maxing pooling in the CNN still outputs an SP (polynomial approximation): we should see the inputs of a max function in NNs as the outputs of several (such as two or four) polynomial functions and regard the output of a max function as another polynomial (SP). A toy example is that the max function for two functions $y = 0 \times x + 0$ and $y = 1 \times x + 0$ is still a function $y = \max(0,x)$ ( ReLU ) that can be approximated by a polynomial. 

\subsection{An Alternative to NNs}
Supposing that we have a dataset $X$ with its observation $Y$. Depending on a regression or a classification mission, we could first select any neural network including the deep-learning ones, and then obtain the SPs (polynomials) by replacing all the activation functions with polynomial approximations. We can actually obtain several SPs based on the given information ($X$ and $Y$). Then by using again the method of undetermined coefficients from mathematics, we can obtain a NN that perfectly completes the mission, whether it is a regression one or a classification one. 

Thus, $\textbf{ the computational complexity of the training algorithm for NNs can, at least, be reduced } $ $\textbf{ to solving a set of multivariate higher-degree equations }$.  The $\texttt{ fsolve }$ function in MatLab can help to solve such problems by using a combination of numerical methods, including the Newton-Raphson method.

\section{Experiments}
In this section, we show how to provide an alternative to NNs for both classification and regression by using the method of undetermined coefficients in mathematics. For simplicity, we use a NN with only one hidden layer of four nodes and select the square function $\Phi(x) = x^2$ as the activation function.

\subsection{Experiment 1} 
\begin{figure}[!ht]
\centering
\includegraphics[scale=1.]{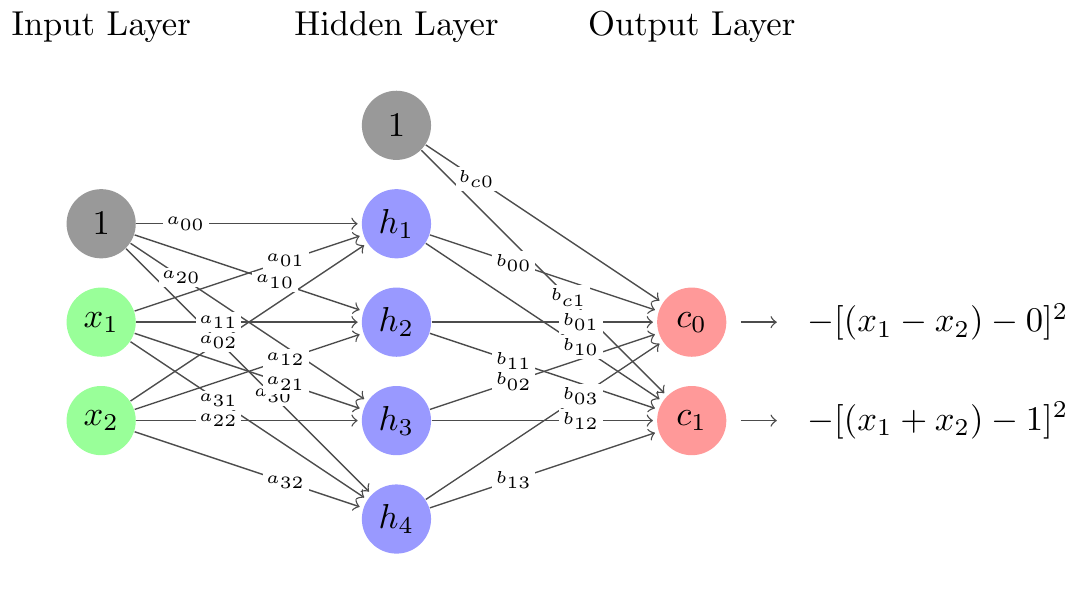}
\caption{
 An Alternative to NNs for classification }
\label{ MyNNinterpret }
\end{figure}

Supposing that we have two classes with labels $c_0 = 0$ and $c_1 = 1$, generating from   $x_1 - x_2 (= 0)$ and $x_1 + x_2 (= 1)$ , respectively, as shown in Figure~\ref{ MyNNinterpret }.

We can first build two polynomials (SPs) for these two classes:  $\bar c_0 = -[ (x_1 - x_2) -0]^2$  and $\bar c_1 = -[ (x_1 + x_2) -1]^2$, respectively:  

\begin{align*}
& c_0 ( = 0) \xmapsto{ } \bar c_0 = -(x_1 - x_2)^2 = -x_1^2 - x_2^2 + 2x_1x_2 ,  \\
& c_1 ( = 1) \xmapsto{ } \bar c_1 = -[ (x_1 + x_2) -1]^2 = -x_1^2 - x_2^2 - 1 -  2x_1x_2  + 2x_1 + 2x_2 .
\end{align*}

\begin{align*}
&
 \begin{bmatrix}
  1        \\ 
  x_{1}    \\
  x_{2}    \\
 \end{bmatrix} ^{\intercal}
 \times
 \begin{bmatrix}
 a_{00}    &   a_{01}   & a_{02}   \\
 a_{10}    &   a_{11}   & a_{12}   \\
 a_{20}    &   a_{21}   & a_{22}   \\
 a_{30}    &   a_{31}   & a_{32}   \\
 \end{bmatrix} ^{\intercal}
 =
 \begin{bmatrix}
 a_{00}    +   a_{01}x_1   + a_{02}x_2       \\
 a_{10}    +   a_{11}x_1   + a_{12}x_2       \\
 a_{20}    +   a_{21}x_1   + a_{22}x_2           \\
 a_{30}    +   a_{31}x_1   + a_{32}x_2       \\
 \end{bmatrix} ^{\intercal}
 \xmapsto{ \Phi(x) = x^2}
\\
& 
 \begin{bmatrix}
 (a_{00}    +   a_{01}x_1   + a_{02}x_2 )^2      \\
 (a_{10}    +   a_{11}x_1   + a_{12}x_2 )^2      \\
 (a_{20}    +   a_{21}x_1   + a_{22}x_2 )^2          \\
 (a_{30}    +   a_{31}x_1   + a_{32}x_2 )^2      \\
 \end{bmatrix} ^{\intercal}
\xmapsto{ } 
 \begin{bmatrix}
 1  \\
 (a_{00}    +   a_{01}x_1   + a_{02}x_2 )^2      \\
 (a_{10}    +   a_{11}x_1   + a_{12}x_2 )^2      \\
 (a_{20}    +   a_{21}x_1   + a_{22}x_2 )^2          \\
 (a_{30}    +   a_{31}x_1   + a_{32}x_2 )^2      \\
 \end{bmatrix} ^{\intercal}
  \times
 \begin{bmatrix}
 b_{c0}    &    b_{c1}      \\
 b_{00}    &    b_{10}      \\
 b_{01}    &    b_{11}      \\
 b_{02}    &    b_{12}      \\
 b_{03}    &    b_{13}      \\
 \end{bmatrix} 
\xmapsto{ }  
 \begin{bmatrix}
 \bar y_{0}    &   \bar y_{1}      \\
 \end{bmatrix} 
\end{align*} 

We deliberately select a simple NN with only four hidden-layer nodes so as to easily calculate its SPs  $\bar y_{0}$ and $\bar y_{1}$:

\begin{align*}
\bar y_{0}  &=  b_{c0}   +   b_{00} (a_{00}    +   a_{01}x_1   + a_{02}x_2 )^2   + b_{01} (a_{10}    +   a_{11}x_1   + a_{12}x_2 )^2     \\  & \hspace{.92cm}  +     b_{02} (a_{20}    +   a_{21}x_1   + a_{22}x_2 )^2          +  b_{03} (a_{30}    +   a_{31}x_1   + a_{32}x_2 )^2     , \\
\bar y_{1}  &=  b_{c1}   +   b_{10} (a_{00}    +   a_{01}x_1   + a_{02}x_2 )^2   + b_{11} (a_{10}    +   a_{11}x_1   + a_{12}x_2 )^2     \\  & \hspace{.92cm}  +     b_{12} (a_{20}    +   a_{21}x_1   + a_{22}x_2 )^2          +  b_{13} (a_{30}    +   a_{31}x_1   + a_{32}x_2 )^2    .  \end{align*}
 
Let $\bar y_{0} = \bar c_0$ and $\bar y_{1} = \bar c_1$. After applying the method of undetermined coefficients from mathematics to these two polynomial equations, we have the following 12 non-linear equations:

\begin{displaymath}
 \left\{ \begin{array}{l}
b_{c0}  +  b_{00}a_{00}a_{00}  +  b_{01}a_{10}a_{10}  +  b_{02}a_{20}a_{20}  +  b_{03}a_{30}a_{30}  =  0     ,  \\
b_{00}a_{01}a_{01}  +  b_{01}a_{11}a_{11}  +  b_{02}a_{21}a_{21}  +  b_{03}a_{31}a_{31}  =  -1              ,  \\
b_{00}a_{02}a_{02}  +  b_{01}a_{12}a_{12}  +  b_{02}a_{22}a_{22}  +  b_{03}a_{32}a_{32}  =  -1              ,  \\
b_{00}a_{00}a_{01}  +  b_{01}a_{10}a_{11}  +  b_{02}a_{20}a_{21}  +  b_{03}a_{30}a_{31}  =  0              ,  \\
b_{00}a_{00}a_{02}  +  b_{01}a_{10}a_{12}  +  b_{02}a_{20}a_{22}  +  b_{03}a_{30}a_{32}  =  0              ,  \\
b_{00}a_{01}a_{02}  +  b_{01}a_{11}a_{12}  +  b_{02}a_{21}a_{22}  +  b_{03}a_{31}a_{32}  =  1              ,  \\
                \\
b_{c1}  +  b_{10}a_{00}a_{00}  +  b_{11}a_{10}a_{10}  +  b_{12}a_{20}a_{20}  +  b_{13}a_{30}a_{30}  =  -1   ,  \\
b_{10}a_{01}a_{01}  +  b_{11}a_{11}a_{11}  +  b_{12}a_{21}a_{21}  +  b_{13}a_{31}a_{31}  =  -1              ,  \\
b_{10}a_{02}a_{02}  +  b_{11}a_{12}a_{12}  +  b_{12}a_{22}a_{22}  +  b_{13}a_{32}a_{32}  =  -1              ,  \\
b_{10}a_{00}a_{01}  +  b_{11}a_{10}a_{11}  +  b_{12}a_{20}a_{21}  +  b_{13}a_{30}a_{31}  =  1              ,  \\
b_{10}a_{00}a_{02}  +  b_{11}a_{10}a_{12}  +  b_{12}a_{20}a_{22}  +  b_{13}a_{30}a_{32}  =  1              ,  \\
b_{10}a_{01}a_{02}  +  b_{11}a_{11}a_{12}  +  b_{12}a_{21}a_{22}  +  b_{13}a_{31}a_{32}  =  -1.        
  \end{array} \right.
\end{displaymath}

Finally, We use the function $\texttt{ fsolve }$ in Octave to solve this  set of non-linear equations  by selecting the ``1'' row vector as the initial guesses for the variable values, obtaining the desired result:

\begin{align*}
& \begin{bmatrix}
 a_{00}    &   a_{01}   & a_{02}   \\
 a_{10}    &   a_{11}   & a_{12}   \\
 a_{20}    &   a_{21}   & a_{22}   \\
 a_{30}    &   a_{31}   & a_{32}   \\
 \end{bmatrix} 
 =
 \begin{bmatrix}
 -0.9642548 &  0.9650999 &  0.9635186   \\
 0.0311467 & -0.1118354 &  0.0675761  \\
 0.0024421 & -1.0605552 &  1.0563675   \\
 0.0170079 &  0.0285231 & -0.0109594   \\
 \end{bmatrix} ,
 \\
 &
  \begin{bmatrix}
 b_{c0}    &    b_{c1}      \\
 b_{00}    &    b_{10}      \\
 b_{01}    &    b_{11}      \\
 b_{02}    &    b_{12}      \\
 b_{03}    &    b_{13}      \\
 \end{bmatrix} 
 =
  \begin{bmatrix}
 -0.00062347   &   -0.00041223     \\
 -0.00055342   &   -1.07563264     \\
 0.93191963    &   0.36561330    \\
 -0.89956744   &   -0.00282642  \\
 0.82785662    &   0.57792885     \\
 \end{bmatrix} .
\end{align*} 
Further testing on this result via applying the ordinary method ( log-likelihood loss function with the softmax function ) for this classification mission shows that the simple NN with weights of this result can predict the right class.

In this way, it is possible to find a NN to perfectly classify the MNIST datasets. We can select the SP $$P1 = - \prod_{  i \textit{ where } y_i = \bar y  } [ \sum_{j = 0}^d (f_j - x_{ij})^2 ] $$ for each image class $\bar y$ and then follow the above method. 
 In this case, the squared activation function had to be replaced with a high-enough degree polynomial perfectly approximating some popular activation function over a large range, say the range $[-1e8, +1e8]$, in order to elevate the input linear space of the NN to the SP $P1$.

\subsection{Experiment 2}
\begin{figure}[!ht]
\centering
\includegraphics[scale=1.]{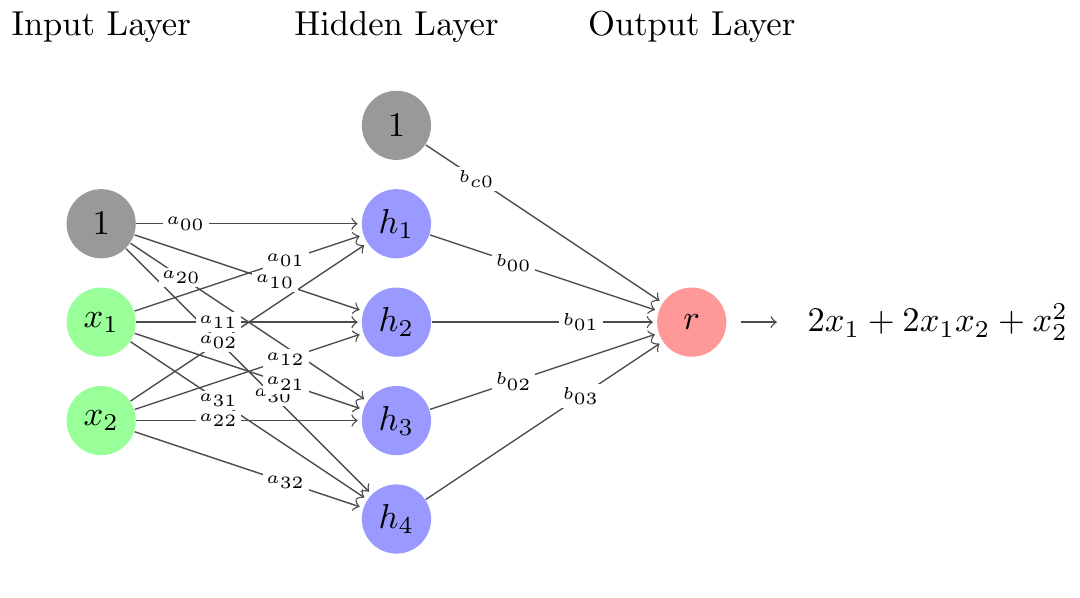}
\caption{
 An Alternative to the NN for regression }
\label{ MyNNinterpretRegression }
\end{figure}

Supposing that we have one regression mission to predict the output of the function $r = 2x_1 + 2x_1x_2 + x_2x_2$, generating from the function  $2x_1 + 2x_1x_2 + x_2x_2 (= r)$ , as shown in Figure~\ref{  MyNNinterpretRegression  }.

We can also build an SP ( a polynomial) for this regression:

$$ \bar r ( = 2x_1 + 2x_1x_2 + x_2x_2) \xmapsto{ SP } 2x_1 + 2x_1x_2 + x_2x_2$$

Like in Experiment $1$, we have the following   set of $6$ non-linear equations:  
\begin{displaymath}
 \left\{ \begin{array}{l}
b_{c0}  +  b_{00}a_{00}a_{00}  +  b_{01}a_{10}a_{10}  +  b_{02}a_{20}a_{20}  +  b_{03}a_{30}a_{30}  =  0             ,  \\
b_{00}a_{01}a_{01}  +  b_{01}a_{11}a_{11}  +  b_{02}a_{21}a_{21}  +  b_{03}a_{31}a_{31}  =  0             ,  \\
b_{00}a_{02}a_{02}  +  b_{01}a_{12}a_{12}  +  b_{02}a_{22}a_{22}  +  b_{03}a_{32}a_{32}  =  1             ,  \\
b_{00}a_{00}a_{01}  +  b_{01}a_{10}a_{11}  +  b_{02}a_{20}a_{21}  +  b_{03}a_{30}a_{31}  =  1             ,  \\
b_{00}a_{00}a_{02}  +  b_{01}a_{10}a_{12}  +  b_{02}a_{20}a_{22}  +  b_{03}a_{30}a_{32}  =  0             ,  \\
b_{00}a_{01}a_{02}  +  b_{01}a_{11}a_{12}  +  b_{02}a_{21}a_{22}  +  b_{03}a_{31}a_{32}  =  1          .
  \end{array} \right.
\end{displaymath}

and obtain the following result:

\begin{align*}
 \begin{bmatrix}
 a_{00}    &   a_{01}   & a_{02}   \\
 a_{10}    &   a_{11}   & a_{12}   \\
 a_{20}    &   a_{21}   & a_{22}   \\
 a_{30}    &   a_{31}   & a_{32}   \\
 \end{bmatrix} 
 =
 \begin{bmatrix}
 -1.54767   &    0.99491    &   3.02228   \\
  0.40528    &   2.19680   &    0.52124  \\
  0.14781    &   2.20493    &   0.14896   \\
  0.95912   &    1.64751    &   0.50298   \\
 \end{bmatrix} ,
 \ \ \ 
  \begin{bmatrix}
 b_{c0}        \\
 b_{00}       \\
 b_{01}        \\
 b_{02}       \\
 b_{03}       \\
 \end{bmatrix} 
 =
  \begin{bmatrix}
 -0.830416     \\
  0.081054    \\
0.430706    \\
 -0.797845  \\
 0.633714    \\
 \end{bmatrix} .
\end{align*}

The Octave code to verify the validity of this result is as follows:

\newpage
\begin{lstlisting}
>> # [1 (X' * A') .* (X' * A')]  * (B') 
>> A = [-1.547668,   0.994909,   3.022282;   0.405276,   2.196802,   0.521244;   0.147810,   2.204926,   0.148959;   0.959119,   1.647513,   0.502981];
>> B = [-0.830416,   0.081054,   0.430706,  -0.797845,   0.633714];
>> X = [1;1;1];
>> XA = X' * A'
XA =

   2.4695   3.1233   2.5017   3.1096

>> XA = XA .* XA
XA =

   6.0985   9.7551   6.2585   9.6697

>> XA = [1 XA]
XA =

   1.0000   6.0985   9.7551   6.2585   9.6697

>> XA * (B')
ans =  5.0000
>> 
>> X = [1;1;1];
>> [1 (X' * A') .* (X' * A')]  * (B')
ans =  5.0000
>> X = [1;2;1];
>> 2 *2 + 2 *2 *1 +1*1
ans =  9
>> [1 (X' * A') .* (X' * A')]  * (B')
ans =  9.0000
>>
\end{lstlisting}

Full Octave code as well as other data is openly available at \href{https://github.com/petitioner/InterpretNN}{$\texttt{https://github.com/petitioner/InterpretNN}$}.

\subsection{Experiment 3}

In this experiment, we creat a fake dataset of size 4 for a toy example  example of classification, as shown in the Table~\ref{tab1}.
First, we  build two SPs from the Table~\ref{tab1}:

\begin{table}[htbp]
\centering
\caption{A fake dataset for a toy example of classification }
\label{tab1}
\begin{tabular}{|l|c|c|c|}
\hline
$f_0$ &  \mysplit{ $f_1$ } & \mysplit{ $f_2$  }   &  Class Labels     \\
\hline
  1    &     0.1     &     0.6    & 3    \\ 
\hline
  1    &     0.2     &     0.7    & 3    \\ 
\hline  
  1    &     0.3     &     0.8    & 8   \\
\hline
  1    &     0.4     &     0.9    & 8   \\
\hline
\end{tabular}
\end{table}

\begin{align*}
 \bar c_0 ( = 0)  \xmapsto{ } SP_0 & = - \prod_{  i \textit{ where } y_i = 3  } [ \sum_{j = 0}^d (f_j - x_{ij})^2 ] \\ & = -[(x_1 - 0.1)^2 + (x_2 - 0.6)^2]\cdot[(x_1 - 0.2)^2 + (x_2 - 0.7)^2]  \\
 & =  -x_1^4 + 0.6 \cdot x_1^3 - 2 \cdot x_1^2 \cdot x_2^2 + 2.6 \cdot x_1^2 \cdot x_2 - 0.98 \cdot x_1^2 + 0.6 \cdot x_1 \cdot x_2^2  \\ & \ \ \ \ \  - 0.76 \cdot x_1 \cdot x_2 + 0.254 \cdot x_1 - x_2^4 + 2.6 \cdot x_2^3 - 2.58 \cdot x_2^2 + 1.154 \cdot x_2 - 0.1961  ,  \\
 \bar c_1 ( = 1)  \xmapsto{ } SP_1 &= - \prod_{  i \textit{ where } y_i = 8  } [ \sum_{j = 0}^d (f_j - x_{ij})^2 ] \\ & =  -[(x_1 - 0.3)^2 + (x_2 - 0.8)^2]\cdot[(x_1 - 0.4)^2 + (x_2 - 0.9)^2]  \\
 & =  -x_1^4 + 1.4 \cdot x_1^3 - 2 \cdot x_1^2 \cdot x_2^2 + 3.4 \cdot x_1^2 \cdot x_2 - 2.18 \cdot x_1^2 + 1.4 \cdot x_1 \cdot x_2^2  \\ &  \ \ \ \ \    - 2.36 \cdot x_1 \cdot x_2    + 1.166 \cdot x_1   - x_2^4 + 3.4 \cdot x_2^3 - 4.58 \cdot x_2^2 + 2.866 \cdot x_2 - 0.7081 .
\end{align*} 

Since $SP_0$ and $SP_1$ are both polynomials of degree 4, we have to replace the degree-2  polynomial activation function with the polynomial of degree 4. We also use a NN with only one-hidden layer of 8 nodes: 
\begin{align*}
&
 \begin{bmatrix}
  1        \\ 
  x_{1}    \\
  x_{2}    \\
 \end{bmatrix} ^{\intercal}
 \times
 \begin{bmatrix}
 a_{00}    &   a_{01}   & a_{02}   \\
 a_{10}    &   a_{11}   & a_{12}   \\
 a_{20}    &   a_{21}   & a_{22}   \\
 a_{30}    &   a_{31}   & a_{32}   \\
 a_{40}    &   a_{41}   & a_{42}   \\
 a_{50}    &   a_{51}   & a_{52}   \\
 a_{60}    &   a_{61}   & a_{62}   \\
 a_{70}    &   a_{71}   & a_{72}   \\
 \end{bmatrix} ^{\intercal}
 =
 \begin{bmatrix}
 a_{00}    +   a_{01}x_1   + a_{02}x_2       \\
 a_{10}    +   a_{11}x_1   + a_{12}x_2       \\
 a_{20}    +   a_{21}x_1   + a_{22}x_2           \\
 a_{30}    +   a_{31}x_1   + a_{32}x_2       \\
 a_{40}    +   a_{41}x_1   + a_{42}x_2       \\
 a_{50}    +   a_{51}x_1   + a_{52}x_2       \\
 a_{60}    +   a_{61}x_1   + a_{62}x_2       \\
 a_{70}    +   a_{71}x_1   + a_{72}x_2       \\
 \end{bmatrix} ^{\intercal}
 \xmapsto{ \Phi(x) = x^4}
\\
& 
 \begin{bmatrix}
 (a_{00}    +   a_{01}x_1   + a_{02}x_2 )^4      \\
 (a_{10}    +   a_{11}x_1   + a_{12}x_2 )^4      \\
 (a_{20}    +   a_{21}x_1   + a_{22}x_2 )^4          \\
 (a_{30}    +   a_{31}x_1   + a_{32}x_2 )^4      \\
 (a_{40}    +   a_{41}x_1   + a_{42}x_2 )^4      \\
 (a_{50}    +   a_{51}x_1   + a_{52}x_2 )^4      \\
 (a_{60}    +   a_{61}x_1   + a_{62}x_2 )^4      \\
 (a_{70}    +   a_{71}x_1   + a_{72}x_2 )^4      \\
 \end{bmatrix} ^{\intercal}
\xmapsto{ } 
 \begin{bmatrix}
 1  \\
 (a_{00}    +   a_{01}x_1   + a_{02}x_2 )^4      \\
 (a_{10}    +   a_{11}x_1   + a_{12}x_2 )^4      \\
 (a_{20}    +   a_{21}x_1   + a_{22}x_2 )^4          \\
 (a_{30}    +   a_{31}x_1   + a_{32}x_2 )^4      \\
 (a_{40}    +   a_{41}x_1   + a_{42}x_2 )^4      \\
 (a_{50}    +   a_{51}x_1   + a_{52}x_2 )^4      \\
 (a_{60}    +   a_{61}x_1   + a_{62}x_2 )^4      \\
 (a_{70}    +   a_{71}x_1   + a_{72}x_2 )^4      \\
 \end{bmatrix} ^{\intercal}
  \times
 \begin{bmatrix}
 b_{c0}    &    b_{c1}      \\
 b_{00}    &    b_{10}      \\
 b_{01}    &    b_{11}      \\
 b_{02}    &    b_{12}      \\
 b_{03}    &    b_{13}      \\
 b_{04}    &    b_{14}      \\
 b_{05}    &    b_{15}      \\
 b_{06}    &    b_{16}      \\
 b_{07}    &    b_{17}      \\
 \end{bmatrix} 
\xmapsto{ }  
 \begin{bmatrix}
 \bar y_{0}    &   \bar y_{1}      \\
 \end{bmatrix} 
\end{align*}

The Octave code to verify the validity of this result is as follows:

\subsection{Experiment 4}

Given a dataset  $X \in \mathbb{R}^{n \times (1 + d)}$ with its predictions  $Y \in \mathbb{N}^{n \times 1}$ for a toy example of regression, we can find a polynomial approximation of the single SP from this dataset:
$$ c_0 \cdot {(a_{0} + a_{01}x_1 + \cdots + a_{0d}x_d)}^{0} + \cdots + c_N \cdot {(a_{0} + a_{01}x_1 + \cdots + a_{0d}x_d)}^{N} , $$
which could be used to generate $n$ nonlinear equations: 

\begin{displaymath}
 \left\{ \begin{array}{l}
c_0 \cdot {(a_{0} + a_{01}x_{11} + \cdots + a_{0d}x_{1d})}^{0} + \cdots + c_N \cdot {(a_{0} + a_{01}x_{11} + \cdots + a_{0d}x_{1d})}^{N}  =  y_1             ,  \\
   \vdots     \\
c_0 \cdot {(a_{0} + a_{01}x_{n1} + \cdots + a_{0d}x_{nd})}^{0} + \cdots + c_N \cdot {(a_{0} + a_{01}x_{n1} + \cdots + a_{0d}x_{nd})}^{N}  =  y_n        .
  \end{array} \right.
\end{displaymath}

Thus, we can also obtain the weights of a specialized NN for a regression mission by solving this set of $n$ nolinear equations.

\newpage
\begin{lstlisting}
>> # [1 (X' * A') .* (X' * A') .* (X' * A') .* (X' * A')]  * (B') 
>> A = [ -0.984418867,  -1.968508208,   1.968508477;  -0.424872463,   0.732323761,   0.195942362; -0.552682441,   0.201900520,   0.749010626;   0.699046565,  -0.000884899,  -0.000884774;  0.212278656,    -0.660869011,  -0.176822978;   0.315521747,  -0.008207178,  -0.166449823;   0.569270235,  -0.171211038,  -0.635132912;  -1.576597281,  -0.034112077,  -3.607466152];
>>
>> B = [ 0.131064245,  -0.014791045,    -0.028456215,  -3.181782656,  -0.508710969,  -4.013970304,    -3.649065294,   1.432325871,  -0.000016396; -0.394862732,  -0.014791104,  -2.662421623,   0.730988408,   1.705192444,  -0.042404801,   -4.392615150,  -6.134780702,  -0.000013827 ];
>>
>> X = [1;   0.1;    0.6];
>> [1 (X' * A') .* (X' * A') .* (X' * A') .* (X' * A')]  * (B')
ans =

  -0.0000000032924  -0.0143999650938

>> X = [1;   0.2;    0.7];
>> [1 (X' * A') .* (X' * A') .* (X' * A') .* (X' * A')]  * (B')
ans =

   0.000000012959  -0.001599893885

>> X = [1;   0.3;    0.8];
>> [1 (X' * A') .* (X' * A') .* (X' * A') .* (X' * A')]  * (B')
ans =

  -0.00159995996   0.00000018751

>> X = [1;   0.4;    0.9];
>> [1 (X' * A') .* (X' * A') .* (X' * A') .* (X' * A')]  * (B')
ans =

  -0.01439992034   0.00000028523

>>
\end{lstlisting}

\section{Conclusion}

In this paper, we presented a new prospect of viewing NNs, as essentially a form of SS. We have shown that the modern activation functions, equivalent to some infinite-degree polynomials, are acting as a magnifying function mapping the raw linear space to an infinite-dimensional space that we term Super Space (SS). This present theory could interpret why NNs can work for both regression and classification tasks --- this is because of applying various Super Planes.

Most importantly, we have shown that the complexity of training NNs including deep-learning ones can be reduced to that of solving a system of non-linear equations via the method of undetermined coefficients from mathematics to obtain the weights of a pre-selected NN architecture. Given a dataset, whether it is for regression or classification, we can determine the weights of any specialized NN even without applying the ordinary solving method used in the training phase.

There is still a lot of work to be done. We hope to include more experiments to prove the theory proposed in this paper. However, the programming task is too heavy for current authors. In addition, Chiang is currently looking for a PhD to restart his research and studies, and he believes that his best chance is to find a position in privacy-preserving machine learning, rather than explaining neural networks. Therefore, this work, now and in the future, is of no help to him. Even so, future work on this paper may still be done.


\bibliography{InterpretNN}
\bibliographystyle{apalike}  

\end{document}